%% file: main.tex
\documentclass[letterpaper]{article} 
\usepackage{array}
\usepackage{aaai25}  
\usepackage{times}  
\usepackage{helvet}  
\usepackage{courier}  
\usepackage[hyphens]{url}  
\usepackage{graphicx} 
\usepackage{natbib}  
\usepackage{caption} 
\usepackage{algorithm}
\usepackage{algorithmic}
\usepackage{amsmath}
\usepackage{amssymb}
\usepackage{amsfonts}
\usepackage{subcaption}

\usepackage{tcolorbox}
\usepackage{bm}
\usepackage{booktabs}
\usepackage{amsthm}
\usepackage{multirow}
\usepackage{framed}

\usepackage{url}
\makeatletter
\newcommand{\figcaption}[1]{\def\@captype{figure}\caption{#1}}
\newcommand{\tblcaption}[1]{\def\@captype{table}\caption{#1}}
\makeatother
\newcount\K
\def\bla#1{
\K=0 \loop\ifnum\K<#1
{\textcolor[gray]{0.9}{{\it bla bla bla bla bla bla bla bla bla bla bla bla bla bla bla}}}
\advance\K by1\repeat
}
\newcommand{\todo}[1]{
\ifx#10
\textcolor{red}{$0.00_{\pm 0.00}$}
\else
\textcolor{red}{#1}
\fi
}

\usepackage{comment}

\usepackage{newfloat}
\usepackage{listings}
\DeclareCaptionStyle{ruled}{labelfont=normalfont,labelsep=colon,strut=off} 
\floatstyle{ruled}
\newfloat{listing}{tb}{lst}{}
\floatname{listing}{Listing}
\usepackage{color}

\title{Rectified Lagrangian for Out-of-Distribution Detection\\in Modern Hopfield Networks}
\author{
Ryo Moriai\thanks{Equal contribution}\textsuperscript{\rm 1},
Nakamasa Inoue\footnotemark[1]\textsuperscript{\rm 1},
Masayuki Tanaka\textsuperscript{\rm 1},
Rei Kawakami\textsuperscript{\rm 1},\\
Satoshi Ikehata\textsuperscript{\rm 1, \rm 3},
Ikuro Sato\textsuperscript{\rm 1,\rm 2}
}
\affiliations{
\textsuperscript{\rm 1} Institute of Science Tokyo, Japan \\
\textsuperscript{\rm 2} Denso IT Laboratory, Inc.~Japan \\
\textsuperscript{\rm 3} National Institute of Informatics, Japan\\
moriai.ryo@d-itlab.titech.ac.jp,
inoue@comp.isct.ac.jp
}

\usepackage{bibentry}
\begin{document}

\maketitle
\input{0_abstract}

%

\input{1_introduction}

\input{2_related_work}

\input{3_modern_hopfield_network}

\input{4_method}

\input{5_training}

\input{6_experiments}

\input{7_conslusion}
\bibliography{main}
\input{8_appendix}

\end{document}

%% file: 0_abstract.tex

\begin{abstract}
Modern Hopfield networks (MHNs) have recently gained significant attention in the field of artificial intelligence because they can store and retrieve a large set of patterns with an exponentially large memory capacity.
A MHN is generally a dynamical system defined with Lagrangians of memory and feature neurons,
where memories associated with in-distribution (ID) samples are represented by attractors in the feature space.
One major problem in existing MHNs lies in managing out-of-distribution (OOD) samples because it was originally assumed that all samples are ID samples.
To address this, we propose the rectified Lagrangian
(RegLag), a new Lagrangian for memory neurons
that explicitly incorporates an attractor for OOD samples in the dynamical system of MHNs.
RecLag creates a trivial point attractor for any interaction matrix, enabling OOD detection by identifying samples that fall into this attractor as OOD. 
The interaction matrix is optimized so that the probability densities can be estimated to identify ID/OOD.
We demonstrate the effectiveness of RecLag-based MHNs compared to energy-based OOD detection methods, including those using state-of-the-art Hopfield energies, across nine image datasets.
\end{abstract}


%% file: 1_introduction.tex
\section{Introduction}

\input{fig_teaser}
Associative memory models have been proposed to model memory retrieval in the brain through fixed-point search in an artificial neural network. Hopfield networks~\citep{hopfield82,hopfield88} are classic examples, based on the idea of using recurrently connected neurons to store and retrieve memory patterns. Although these models are theoretically sound, they suffer limited memory capacity, as the number of distinct memory patterns is at most proportional to the dimension of the feature space.
Recently, numerous studies have explored models with significantly increased memory capacity, the so-called modern Hopfield networks (MHNs)~\citep{krotov16dense,demircigil17hugecapacity,krotov2018dense,BARRA2018205,agliari2020tolerance}.
Some of them are known to have an exponentially large memory capacity with respect to the feature dimension~\cite{demircigil17hugecapacity}.

From a theoretical perspective,
\citet{krotov2021large} introduced a dynamical system that represents associative memory in a continuous time space based on two-body interactions between neurons.
In their system, two Lagrangian functions, one for memory neurons and one for feature neurons, determine the model dynamics.
When certain pairs of Lagrangian functions are chosen,
the system is reduced to classical Hopfield networks~\cite{hopfield82},
dense associative memory~\cite{krotov16dense, demircigil17hugecapacity},
or the MHNs described in \citet{ramsauer21hopfieldallyouneed}, indicating that new Lagrangian function designs could lead to new MHNs.

While these studies have expanded the potential of MHNs both theoretically and practically, one of the primary limitations of existing MHNs lies in managing out-of-distribution (OOD) samples.
The dynamical system does find a fixed point for any test input; \textit{i.e.}, an OOD sample is inevitably associated with one of the memorized in-distribution (ID) samples.
\citet{zhang23she} proposed an OOD-sample detection method based on the Hopfield energy function.
However, they lack theoretical foundation explaining the relationship between the energy and the probability of the input/transient states.
We are thus motivated to develop new MHNs equipping probability-aware OOD rejection functionality within the fixed-point search mechanism.

In this paper, we propose the rectified Lagrangian (RegLag), a new Lagrangian for memory neurons that creates an attractor for OOD samples in the dynamical system of MHNs, as shown in Fig.~\ref{fig:teaser}.
RegLag introduces a rectified linear unit (ReLU) with a constant indicating the ID memory strength to the Lagrangian function of memory neurons.
We theoretically show that 1) RecLag creates a trivial point attractor for any interaction matrix and 2) RecLag-based MHNs are reduced to vanilla MHNs when the ID memory strength is infinitely large, indicating our approach is a natural extension of existing MHNs.
We further devise a training method for RecLag-based MHNs via probabilistic interaction, along with a probability density estimated for ID samples by optimizing the interaction matrix.
Our contributions are summarized below:
\begin{enumerate}
\item We propose RecLag, a new Lagrangian
for memory neurons. RecLag is designed to create a trivial point attractor for any interaction matrix, enabling OOD detection by identifying samples that fall into the attractor as OOD.
\item We propose a training method for RecLag-based MHNs having a probabilistic interaction between memory and feature neurons. We prove that samples with low probability fall into the special attractor created by RecLag.
\item We demonstrated the effectiveness of our approach in comparison with energy-based OOD detection methods, including those using state-of-the-art Hopfield energy functions~\cite{zhang23she} on nine image datasets.
\end{enumerate}

%% file: fig_teaser.tex
\begin{figure}
\centering
\vspace{10pt}
\includegraphics[width=\linewidth]{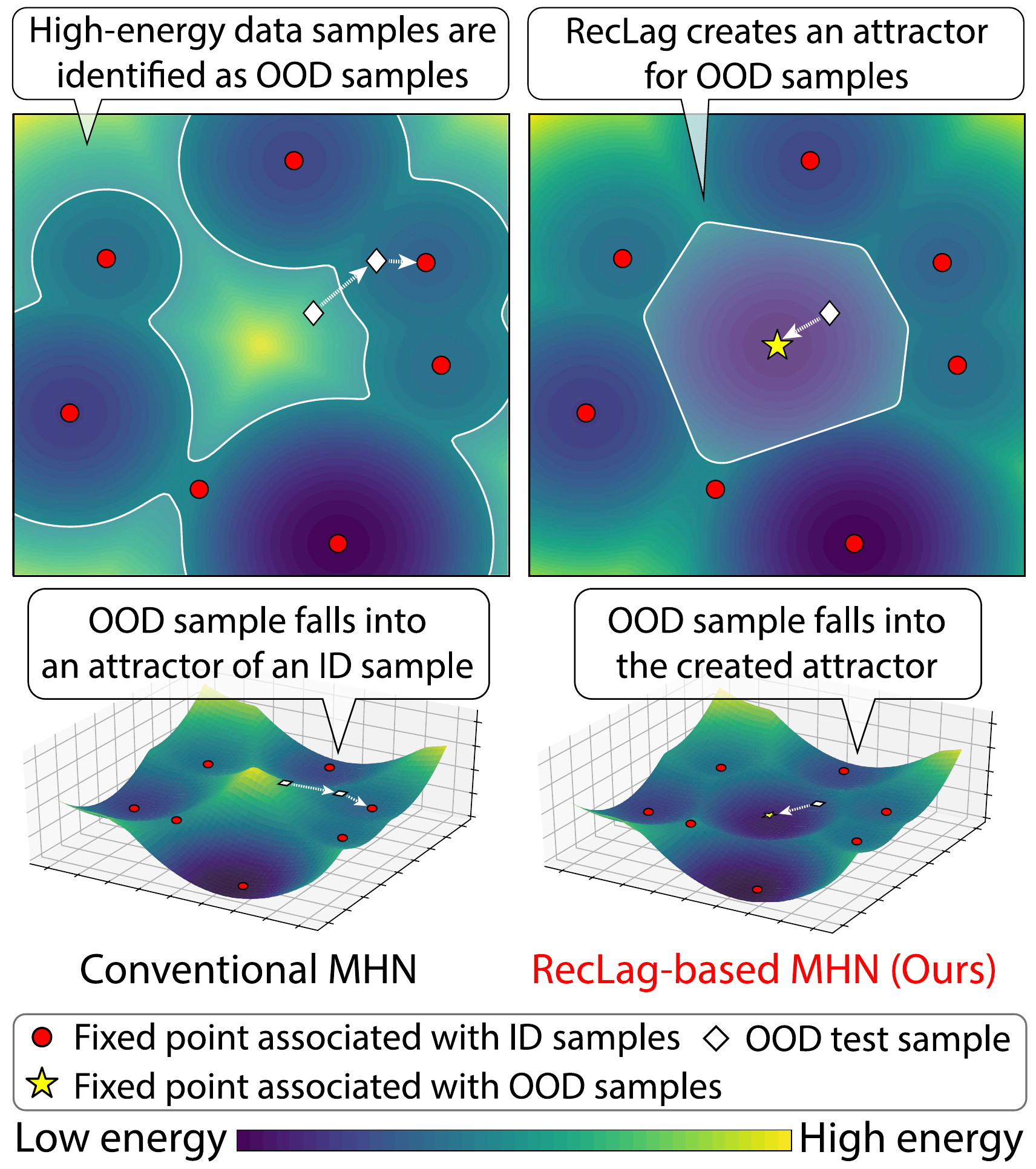}
\caption{
Visualization of energy landscapes. Existing methods identify samples with high Hopfield energy as OOD, though such samples fall into the attractors associated with ID samples.
In contrast, our RecLag-based MHNs possess a dedicated attractor that specifically captures OOD samples.
}
\label{fig:teaser}
\end{figure}

%% file: 2_related_work.tex
\section{Related Work}

\noindent
\textbf{Hopfield Networks.}
Hopfield networks~\cite{hopfield82, hopfield88} are a type of artificial neural network with recurrent structures that model associative memory.
Their development laid the foundation for later models such as Boltzmann machines~\cite{Ackley1985boltzmannmachines} and long short-term memory~\cite{lstm} in the latter part of the 20th century.

In recent years, MHNs, also known as dense associative memory~\cite{krotov16dense}, have been attracting attention because they can have an exponentially large memory capacity \cite{demircigil17hugecapacity}.
Numerous studies have demonstrated the effectiveness of MHNs on various tasks including image classification~\cite{furst2021cloob, ota2023learning}, immune repertoire classification~\cite{widrich20modern},
tabular data classification~\cite{schafl2021hopular},
reaction template prediction~\cite{seidl2022improving},
predictive coding~\cite{NEURIPS2021_1fb36c4c} and reinforcement learning~\cite{widrich2021modern}.

MHNs are formulated as dynamical systems described by analytical differential equations.
Specifically, \citet{ramsauer21hopfieldallyouneed} generalized the energy function from discrete states to continuous states, and then \citet{krotov2021large} formulated the dynamical system of MHNs with two-body differential equations.
Follow-up studies, such as work on universal Hopfield networks~\cite{millidge2022universal}, have further generalized the dynamical system.

\noindent \textbf{OOD Detection.}
OOD detection aims to identify data samples that deviate from the distribution of training data samples.
This paper focuses on post hoc approaches, where the detection mechanism is applied after the model has been trained.
One of the most well-known approaches is maximum softmax probability (MSP) scoring~\cite{hendrycks2017msp}, which uses the highest softmax output score to identify OOD samples, under the assumption that ID samples yield higher MSP scores compared to OOD samples.
To more precisely estimate the distribution of OOD samples, various enhancements and alternative post hoc methods have been proposed~\cite{liang2018odin,liu2020energy,sun2021react,sun2022dice,shen2023posthoc,chen2024tagfog}.
Among them, energy-based OOD detection approaches~\cite{liu2020energy, sun2021react} are related to this study in the sense that MHNs have a scalar-valued function associated with the network states, the so-called the Hopfield energy.

Most recently, several pioneering studies have demonstrated the effectiveness of Hopfield energy for OOD detection~\cite{zhang23she, Hofmann2024boosting}.
Their methods identify data samples with high Hopfield energy as OOD samples and achieve superior performance among energy-based OOD detection methods.
However, from a theoretical perspective, every test sample, including an OOD sample, falls into one of the attractors representing a memory pattern associated with an ID data sample as the dynamical system of MHNs evolves over time.
To address this problem, this paper explores MHNs that explicitly have an attractor for OOD samples.

%% file: 3_modern_hopfield_network.tex
\newcommand{\Lagh}{L_{H}}
\newcommand{\Lagv}{L_{V}}
\newcommand{\Numh}{N_{H}}
\newcommand{\Numv}{N_{V}}
\newcommand{\tauh}{\tau_{H}}
\newcommand{\tauv}{\tau_{V}}

\section{Modern Hopfield Networks}

\subsection{Lagrangian-Based Dynamical System}
\textbf{Notation and Settings.}
This paper discusses MHNs with the Lagrangian-based dynamical system proposed by \citet{krotov2021large}.
We denote the feature neurons as $v(t) \in \mathbb{R}^{\Numv}$ and the memory neurons as $h(t) \in \mathbb{R}^{\Numh}$, both at continuous time $t \in \mathbb{R}_{\ge 0}$, where $\Numv, \Numh \in \mathbb{N}$ are the numbers of neurons.
The dynamical system is described by the following differential equations:
\begin{align}
\label{eq:de_v}
\tauv \frac{d v_{i}(t)}{d t} &=\sum_{\mu=1}^{\Numh} \xi_{i \mu} f_{\mu}(h(t)) - v_{i}(t),  \\
\label{eq:de_h}
\tauh \frac{d h_\mu(t)}{d t} &=\sum_{i=1}^{\Numv} \xi_{\mu i} g_{i}(v(t)) - h_{\mu}(t),
\end{align}
where $\xi \in \mathbb{R}^{\Numh \times \Numv}$ is an interaction matrix representing the strength of synapses,
$f: \mathbb{R}^{\Numh} \to \mathbb{R}^{\Numh}$ and 
$g: \mathbb{R}^{\Numv} \to \mathbb{R}^{\Numv}$ are activation functions,
and $\tauv, \tauh \in \mathbb{R}$ are constants that determine the dynamics of neurons.
The activation functions are determined by the Lagrangians $\Lagh: \mathbb{R}^{\Numh}\to \mathbb{R}$ and $\Lagv: \mathbb{R}^{\Numv}\to \mathbb{R}$ such that
\begin{align}
f(h) = \frac{\partial \Lagh(h)}{\partial h}, \quad g(v) = \frac{\partial \Lagv(v)}{\partial v},
\end{align}
where $h \in \mathbb{R}^{\Numh}$ and $v \in \mathbb{R}^{\Numv}$.
The energy function is then given by
\begin{align}
E(v, h)
=&
\sum_{i=1}^{\Numv} v_{i} g_{i}(v)
-
\Lagv(v)
+
\sum_{\mu=1}^{\Numh} h_{\mu} f_{\mu}(h)\nonumber \\
&-
\Lagh(h)
-
\sum_{\mu, i} f_{\mu}(h) \xi_{\mu i} g_{i}(v).
\end{align}
Note that this energy monotonically decreases; that is, we have
$dE(v(t),h(t))/dt \leq 0$ along the trajectory of the dynamical system when the Hessian matrices of the Lagrangians are positive semi-definite.

\noindent \textbf{Lagrangians.}
If we suppose a fixed interaction matrix $\xi$, then the model dynamics are defined by the choice of the Lagrangians.
For example, when the Lagrangian functions are given by the additive functions
\begin{align}
\label{eq:lag_classic}
\Lagh(h) = \sum_{\mu = 1}^{\Numh} \sigma(h_{\mu}),
\quad
\Lagv(v) = \sum_{i = 1}^{\Numv} |v_{i}|,
\end{align}
where $\sigma: \mathbb{R} \to \mathbb{R}$ is a nonlinear function, the energy function reduces to
\begin{align}
E(v) = 
- \sum_{\mu=1}^{\Numh} \sigma \left( \sum_{i=1}^{\Numv} \xi_{\mu i} \cdot \mathrm{sgn}(v_{i}) \right).
\end{align}
under the adiabatic limit $\tauv \gg \tauh$ when $\xi$ is a symmetric matrix.
This energy function is identical to that of dense associative memory~\citep{krotov16dense}.
Further, when $\sigma(x) = x^{2}$, it reduces to the energy function of the classical Hopfield network~\citep{hopfield82}.

Recently, \citet{krotov2021large} introduced the following Lagrangians:
\begin{align}
\label{eq:lag_modern}
\hspace{-5pt}
\Lagh(h) = \frac{1}{\beta}\text{log} \left(\sum^{\Numh}_{\mu=1} \exp\left(\beta h_{\mu}\right)\right),
\Lagv(v) = \frac{1}{2} \sum^{\Numv}_{i=1} v_{i}^2,
\hspace{-5pt}
\end{align}
where $\beta \in \mathbb{R}_{\ge 0}$ is a constant.
Under the adiabatic limit and when $\beta = 1$, the energy function reduces to
\begin{align}
\label{eq:energy_modern}
E(v) = 
- \log \left( \sum_{\mu=1}^{\Numh} \exp \left( \sum_{i=1}^{\Numv} \xi_{\mu i} v_{i} \right)\right)
+ \frac{1}{2} \sum_{i=1}^{\Numv} v_{i}^{2}.
\end{align}
This energy function is identical to that of the MHNs proposed by \citet{ramsauer21hopfieldallyouneed}.

\subsection{Energy-Based OOD Detection}
Let us consider classification problems and denote the number of ID classes for training as $C$.
The goal of OOD detection is to identify data samples that do not belong to any of the $C$ classes.
\citet{zhang23she} proposed using the energy function of MHNs for OOD detection.
Specifically, they introduced two energy functions: modern Hopfield energy (MHE) and simplified Hopfield energy (SHE).
MHE is obtained by replacing the interaction matrix $\xi$ in Eq.~(\ref{eq:energy_modern}) with a class-specific pattern matrix $S^{c} \in \mathbb{R}^{d \times N}$ and by omitting the second term as follows:
\begin{align}
\mathrm{MHE}(\tilde{v}) = 
- \log \left( \sum_{\mu=1}^{d} \exp \left( \sum_{i=1}^{N} S^{c}_{\mu i} \tilde{v}_{i} \right)\right),
\end{align}
where $\tilde{v} \in \mathbb{R}^{d}$ is a test pattern,
$c \in \{1, 2, \cdots, C\}$ is the classification result of $\tilde{v}$ obtained from a pre-trained classification model,
$d$ is the hidden dimension, and $N$ is the number of stored patterns.
SHE is a Taylor approximation of MHE, but is more effective than MHE at detecting OOD.
It is defined as
\begin{align}
\label{eq:she}
\mathrm{SHE}(\tilde{v}) = 
\frac{1}{d} \sum_{\mu = 1}^{d} \sum_{i = 1}^{N} S_{\mu i}^{c}\tilde{v}_{i}.
\end{align}
OOD samples can be detected by applying a threshold to these energy functions.
However, as the dynamical system of MHNs evolves over time, every test sample falls into an attractor associated with an ID data sample, indicating a lack of theoretical consistency.

%% file: 4_method.tex
\section{Rectified Lagrangian}

This section introduces RecLag, a Lagrangian function that creates a point attractor for OOD samples in the dynamical system of HMNs.
As shown in Figure~\ref{fig:energy},
RecLag creates a point attractor in the feature space.
This attractor is designed to exist for any interaction matrix $\xi$, enabling OOD detection by identifying data samples that fall into it as OOD.

\subsection{Definition}

To incorporate a point attractor for OOD samples in the dynamical system, we propose a minimal yet effective modification to the Lagrangian function of memory neurons.
Specifically, we introduce an {\it inverse memory strength constant} $\gamma$, which determines the strength of ID samples stored in memory, with a $\max$ function to screen out negative values, which is applied in the same way as ReLU.
The proposed RecLag is defined as follows.

\vspace{5pt}
\noindent \textit{\textbf{Definition 1.}
We define RecLag as
\begin{align}
\label{eq:reclag}
\Lagh(h) = \max \left(
\frac{1}{\beta}
\log \left( \frac{1}{\gamma} \sum^{\Numh}_{\mu=1} \exp \left({\beta h_{\mu}}\right) \right), 0 \right),
\end{align}
where $\beta, \gamma \in \mathbb{R}_{\ge 0}$ are constants.
}

\subsection{Existence of a Trivial Point Attractor}
With the dynamical system using RecLag $L_{H}$ in Eq.~(\ref{eq:reclag}) for memory neurons and the Lagrangian $\Lagv$ in Eq.~(\ref{eq:lag_modern}) for feature neurons, Theorem 1 shows that there exists a trivial point attractor at the origin of the feature space for any interaction matrix.

\vspace{5pt}
\noindent \textit{\textbf{Theorem 1.}
Suppose that activation functions $f$ and $g$ in the dynamical system of Eqs.~(\ref{eq:de_v},\ref{eq:de_h}) are given by the derivatives of RecLag $\Lagh$ in Eq.~(\ref{eq:reclag}) and the Lagrangian $\Lagv$ in Eq.~(\ref{eq:lag_modern}), respectively.
For any interaction matrix $\xi \in \mathbb{R}^{\Numh \times \Numv}$,
a trivial point attracting set $A = \{\bm{0}\}$ exists at the origin
$\bm{0} \in \mathbb{R}^{\Numv}$ in the feature space when $\gamma > \Numh$ under the adiabatic limit $\tauv = dt$.}
\vspace{-4pt}
\renewcommand{\proofname}{Sketch of proof}
\begin{proof}
With RecLag, writing the differential equations of the dynamical system in finite differences with $\frac{dv_{i}}{dt} \simeq \frac{v^{(k+1)}_{i} - v^{(k)}_{i}}{\Delta t}$ and $\tauv = \Delta t$
gives the following update rule for feature neurons:
\begin{align}
v^{(k+1)}_{i}
\hspace{-2pt}
=
\hspace{-1pt}
\chi
\hspace{-2pt}
\left(
\hspace{-1pt}
G(v^{(k)})
\hspace{-1pt}
\right)
\hspace{-2pt}
\sum_{\mu=1}^{\Numh}
\xi_{i\mu}
\mathrm{softmax}
\hspace{-2pt}
\left(
\hspace{-2pt}
\beta \sum_{j=1}^{\Numv} 
\xi_{\mu j} v^{(k)}_{j}
\hspace{-3pt}
\right)\hspace{-3pt},\hspace{-3pt}
\end{align}
where $k \in \mathbb{N}$ is a discrete time step, and 
\begin{align}
G(v) &=
\log
\left(
\frac{1}{\gamma}
\sum_{\mu = 1}^{\Numh}
\exp\left({\beta \sum_{j=1}^{\Numv} \xi_{\mu j} v_{j}} \right)
\right),\\
\chi (x) &=
\begin{cases}
1 & (x \geq 0)\\
0 & (x < 0)
\end{cases}.
\end{align}
When $v^{(k)} = \bm{0}$, we have $\chi(G(v^{(k)})) = 0$, and thus we have $v^{(k+1)} = \bm{0}$.
This shows that $\bm{0} \in \mathbb{R}^{\Numv}$ is a fixed point of the dynamical system in the feature space.
Further, with the epsilon neighborhood of the origin $U_{\epsilon} = \{u : \| u \|_{2} < \epsilon\}$, we have $\chi(G(u)) = 0$ for every $u \in U_{\epsilon}$ if $\epsilon$ is small enough. This shows that $A = \{\bm{0}\}$ is an attracting set for every fixed interaction matrix $\xi$.
A full proof is given in Appendix A.
\end{proof}

\subsection{Reduction to Vanilla MHNs}
Along with the existence of the trivial point attractor, it is also worth noting the limit where it disappears.
Theorem 2 shows that RecLag-based MHNs reduce to vanilla MHNs when the memory strength of ID samples is infinitely large, that is, when the inverse memory strength constant $\gamma \to 0$.
This theoretical result indicates that our approach is a natural extension of MHNs.
A proof is given in Appendix B.

\vspace{5pt}
\noindent \textit{\textbf{Theorem 2.}
Let $v_{\mathrm{A}}$ and $v_{\mathrm{B}}$ be feature neurons of a vanilla MHN and a RecLag-based MHN, respectively. Suppose $v_{\mathrm{A}}^{(0)} = v_{\mathrm{B}}^{(0)}$. For every $\epsilon > 0$,
there exists a small $\gamma > 0$ such that $\sup_{k} \|v_{\mathrm{A}}^{(k)} - v_{\mathrm{B}}^{(k)}\|_{2} < \epsilon$.
}

\subsection{Visualization and Discussion}
\begin{figure*}
\centering
\includegraphics[width=1.0\linewidth]{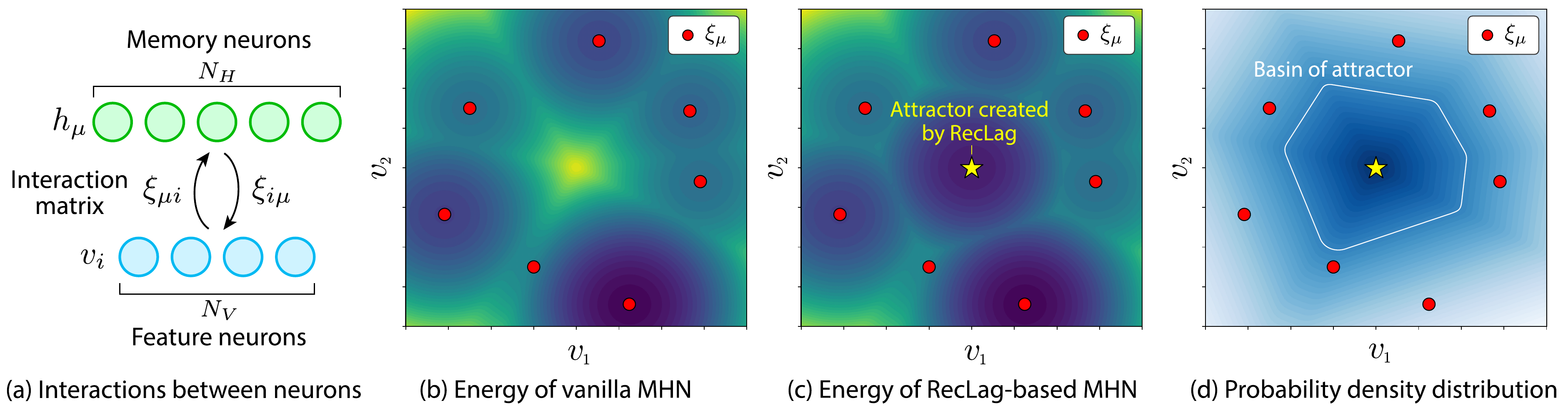}
\caption{
(a) MHN with an interaction matrix $\xi_{\mu i}$ between memory neurons $h_{\mu}$ and feature neurons $v_{i}$.
(b) Energy distribution of a vanilla MHN using the Lagrangians in Eq.~(\ref{eq:lag_modern}).
(c) Energy distribution of a RecLag-based MNH. The point attractor in Theorem 1 created by RecLag is marked by the yellow star.
(d) Probability density distribution in
Eq.~(\ref{eq:reclagprob}).
Data samples with low probability density values fall into the created attractor.
}
\label{fig:energy}
\end{figure*}

\noindent \textbf{Visualization.}
Figure~\ref{fig:energy} compares the energy distributions of a vanilla MHN and a RecLag-based MHN, where each red point indicates a fixed point $\xi_{\mu} \in \mathbb{R}^{N_{V}}$ at a local minimum of the energy function.
As shown, RecLag creates an attractor at the origin of the feature space. This attractor is associated with OOD samples as described in the next section.
The 3D visualization of these energy functions is shown in Figure~\ref{fig:teaser} with trajectories of a test sample (white diamond-shaped point) over time.
As shown, with the vanilla MHN, the test sample falls into one of the attractors even if it is an OOD sample. 
In contrast, with the RecLag-based MHN, the same test sample falls into the created attractor, indicating that none of the memory patterns are associated with it. This shows that the RecLag-based MHN can explicitly manage OOD samples in the dynamical system.

\noindent \textbf{Memory Strength.}
The size of the created attractor increases as the inverse memory strength constant $\gamma$ increases. Consequently, the number of samples identified as OOD samples also increases with $\gamma$.
This indicates that $\gamma$ can serve as a threshold parameter that adjusts the sensitivity of RecLag-based MHNs to OOD samples.
In practice, to draw a receiver operating characteristic (ROC) curve, one could vary $\gamma$ to generate different true positive rates (TPRs) and false positive rates (FPRs) for OOD detection.

%% file: 5_training.tex
\section{Training via Probabilistic Interaction}

This section discuss the basin of the attractor created by RecLag, and proposes a method for training the interaction matrix with ID samples.
Because the basin obviously involves $B_{0} = \{v: G(v) < 0\}$, as shown in the sketch of the proof for Theorem 1, we introduce a method to train the interaction matrix via probabilistic interaction, by which data samples with low probability density values fall into $B_{0}$.

\subsection{Probabilistic Interaction}
The probabilistic interaction explicitly chooses a single memory neuron for each input feature during training in a probabilistic manner.
This creates a cycle of interaction between feature neurons and memory neurons in the following two steps. First, given an input feature $x \in \mathbb{R}^{\Numv}$,
a memory neuron is sampled as $\mu \sim p_{H}(\mu|x)$, where
$\mu \in \{1, 2, \cdots, \Numh\}$ is an index of memory neurons and $p_{H}(\mu|x)$ is a pre-defined conditional probability mass distribution.
Second, given an index $\mu$,
an output feature $y \in \mathbb{R}^{\Numv}$ is sampled as $y \sim p_{V}(y|\mu)$, where $p_{V}(y|\mu)$ is a pre-defined conditional probability density distribution.

Because this interaction can be understood as the stochastic feedforward neural network (SFNN) proposed by \citet{tang13sfnn}, which samples an index of neurons in a hidden layer, we train the interaction matrix using the training method for SFNN.
Specifically, given a set of ID data samples $\mathcal{D} \subset \mathbb{R}^{\Numv}$, the interaction matrix $\xi$ is trained to maximize the sum of probability products:
\begin{align}
\label{eq:objective_sfnn}
P = \sum_{x \in \mathcal{D}} p_{V}(x|\mu) p_{H}(\mu|x).  
\end{align}
Here, the distribution $p_{H} (\mu|x)$ is computed through the joint probability distribution described in the next subsection.
The distribution $p_{V}(x|\mu)$ is used only for training, and thus we use a Gaussian distribution following \citet{tang13sfnn}:\begin{align}
p_{V}(x|\mu)
\hspace{-2pt}
=\hspace{-2pt}
\frac{1}{\sqrt{(2\pi)^{\Numv} |\Sigma|}}
\exp
\hspace{-2pt}
\left(
\hspace{-2pt}
-\frac{1}{2} (x - \xi_{\mu})^{\top} \Sigma^{-1} (x - \xi_{\mu}) \right)
\end{align}
where $\Sigma$ is a learnable covariance matrix.

\subsection{Attracting Probability}
Interestingly, there exists a joint probability distribution $p_{H}(x, \mu)$ that relates the SFNN and the basin $B_{0}$.
Specifically, Definition 2 provides the joint probability distribution, by which the conditional probability for the SFNN is computed as $p_{H}(\mu|x) = p_{H}(x, \mu)/\sum_{\mu}p_{H}(x, \mu)$, and data samples with low probability density values fall into the basin.

\vspace{5pt}
\noindent \textit{\textbf{Definition 2.}
Let $X$ be a continuous random variable of feature neurons over $\mathbb{R}^{\Numv}$, and let $M$ be a discrete random variable of the index of hidden neurons over $\{1, 2, \cdots, \Numh\}$.
We define the joint probability distribution function as
\begin{align}
\label{eq:reclagprob}
p_{H}(X=x, M=\mu) &= \frac{1}{Z} \exp \left(
{\beta \sum_{j=1}^{\Numv} \xi_{\mu j} x_{j}}
\right).
\end{align}
Here, $Z$ is a normalization constant given by
\begin{align}
Z &= \sum_{\mu=1}^{\Numh} \int_{\mathcal{S}}
\exp \left(
{\beta \sum_{j=1}^{\Numv} \xi_{\mu j} x_{j}} \right) dx,
\end{align}
where $\mathcal{S} \in \mathbb{R}^{\Numv}$ is a sufficiently large hypersphere to cover all data samples.
}

\subsection{OOD detection}
Finally, Theorem~3 shows that the probability density distribution $p_{H}(x) = \sum_{\mu} p_{H}(x, \mu)$ explicitly models the distribution of ID samples and that all data samples with a probability density lower than $\delta$ fall into the attractor created by RecLag.
Therefore, OOD samples can be detected by evaluating $p_{H}(\tilde{v})$ given a test sample $\tilde{v} \in \mathbb{R}^{\Numv}$. A proof is given in Appendix C.

\vspace{5pt}
\noindent \textit{\textbf{Theorem 3.}
The basin $B_{0} = \{ v : G(v) < 0\}$ is identical to the set of points that have low probability density values.
In other words, a threshold $\delta$ exists such that
\begin{align}
B_{0} = \{x : p_{H}(X = x) < \delta\}.
\end{align}
}
\vspace{-8pt}

\noindent \textbf{Visualization.}
Figure~\ref{fig:energy}(d) shows the probability density function, where the basin boundary is drawn in white.

\begin{table*}[t]
\centering
\setlength{\tabcolsep}{6.5pt}
{\footnotesize
\begin{tabular}{p{0cm}l|rrrrrrrrr|r}
\toprule
\multicolumn{2}{l|}{Method} & SVHN & LSUN-C & LSUN-R & iSUN & Places & DTD & TIN & SUN & iNaturalist & Average \\
\midrule
\multirow{6}{*}{\rotatebox[origin=c]{90}{ ResNet18}}
 & MSP & 76.34 & 27.52 & 36.54 & 34.84  & 20.55 & 30.65 & 45.82 & 22.89 & 12.62 & 34.19\\
 & Energy & 56.05 & 8.10 & 11.60 & 9.10 & 3.18 & 16.98 & 25.47 & 3.27 & 3.47 & 15.25  \\
 & ReAct & 59.47 & 7.57 & 12.52 & 10.13 & 2.93 & 16.86 & 27.61 & 3.27 & 3.80 & 16.02 \\
 & MHE & 17.59 & 9.20 & 7.68 & 4.74 & 0.33 & 8.96 & 15.86 & \textbf{0.00} & 2.35 & 7.41 \\
 & SHE & \textbf{17.45} & 9.22 & 7.69 & 4.77 & 0.33 & 8.99 & 15.84 & \textbf{0.00} & 2.38 & 7.41 \\ 
 & RecLag & 18.12 & \textbf{6.40}& \textbf{4.60} & \textbf{2.67} & \textbf{0.28} & \textbf{6.82} & \textbf{12.09} & \textbf{0.00} & \textbf{1.68} & \textbf{5.85} \\
 & & $\pm$ 2.02 & \textbf{$\pm$ 0.25}& \textbf{$\pm$ 0.12} & \textbf{$\pm$ 0.47} & \textbf{$\pm$ 0.02} & \textbf{$\pm$ 0.13} & \textbf{$\pm$ 0.25} & \textbf{$\pm $0.00} & \textbf{$\pm$ 0.04} & \textbf{$\pm$ 0.24} \\

\midrule
\multirow{6}{*}{\rotatebox[origin=c]{90}{ ResNet34}}
& MSP & 59.86 & 28.26 & 32.06 & 31.69 & 33.61 & 43.28 & 45.56 & 32.43 & 32.95 & 37.74\\
& Energy& 30.51 & 6.84 & 9.43 & 8.47 & 9.32 & 23.74 & 25.16 & 8.99 & 10.86 & 14.81\\
& ReAct& 45.86 & 14.37 & 14.09 & 13.28 & 15.83 & 29.73 & 31.60 & 15.53 & 11.98 & 21.36\\
& MHE& 6.20 & 6.17 & 4.40 & 2.94 & 2.34 & 14.32 & 15.86 & 0.54 & 4.91 & 6.41\\
& SHE & 6.14 & 6.20 & 4.45 & 3.01 & 2.36 & 14.32 & 15.93 & 0.54 & 4.92 & 6.43 \\
& RecLag & \textbf{5.19} & \textbf{5.60} & \textbf{2.85} & \textbf{2.11} & \textbf{2.31} & \textbf{12.04} & \textbf{11.71} & \textbf{0.33} & \textbf{4.14} & \textbf{5.14} \\
& & \textbf{$\pm$ 0.24} & \textbf{$\pm$ 0.07} & \textbf{$\pm$ 0.05} & \textbf{$\pm$ 0.05} & \textbf{$\pm$ 0.03} & \textbf{$\pm$ 0.07} & \textbf{$\pm$ 0.23} & \textbf{$\pm$ 0.11} & \textbf{$\pm$ 0.08} & \textbf{$\pm$ 0.08} \\

\midrule
\multirow{6}{*}{\rotatebox[origin=c]{90}{WRN40-2}}
& MSP & 41.52 & 44.43 & 38.47 & 39.70 & 33.84 & 35.80 & 51.52 & 34.88 & 27.69 & 38.65\\
& Energy & 15.35 & 17.77 & 14.98 & 17.45 & 10.58 & 19.71 & 36.75 & 9.54 & 8.95 & 16.79\\
& ReAct& 18.83 & 19.93 & 18.25 & 20.68 & 11.98 & 21.67 & 42.02 & 11.44 & 13.26 & 19.78\\
& MHE & 5.40 & 14.60 & 12.03  & 11.48 & 2.90 & 10.99 & 27.28 & \textbf{0.82} & 1.83 & 9.70\\
& SHE & \textbf{5.25} & 14.39 & 13.18  & 12.39 & 2.83 & 10.98 & 28.35 & \textbf{0.82} & 1.84 & 10.00\\
& RecLag & 5.75 & \textbf{7.37} & \textbf{8.44 } & \textbf{8.01 } &  \textbf{2.63}& \textbf{9.75} & \textbf{22.62} & 1.06 & \textbf{1.67}  & \textbf{7.47} \\
& & $\pm$ 0.12& \textbf{$\pm$ 0.18} & \textbf{$\pm$ 0.17} & \textbf{$\pm$ 0.15} & \textbf{$\pm$ 0.05} & \textbf{ $\pm$ 0.10} & \textbf{$\pm$ 0.34} & $\pm$ 0.09 & \textbf{$\pm$ 0.05} &  \textbf{$\pm$ 0.85} \\

\bottomrule
\end{tabular}
}
\caption{OOD detection performance as
FPR95(\%) $\downarrow$ with CIFAR-10 images being ID samples. Our RecLag-based MHN (RecLag) is compared with MSP~\cite{hendrycks2017msp}, Energy~\cite{liu2020energy}, ReAct~\cite{sun2021react}, MHE~\cite{zhang23she}, and SHE~\cite{zhang23she}. 
For the proposed RecLag the trimmed means and standard deviations (following ± symbols) over 11 trials with the largest and the smallest ones being trimmed  are reported.}
\label{tab:fpr}
\end{table*}

%% file: 6_experiments.tex
\section{Experiments}

We focus on evaluating OOD detection performance of our proposed method along with strong baselines in this work.

\subsection{Experimental Settings}


\noindent \textbf{Datasets.}
Eleven image datasets were used to conduct OOD detection experiments:
CIFAR-10~\cite{krizhevsky2009cifar}, CIFAR-100~\cite{krizhevsky2009cifar},
SVHN~\cite{netzer2011svhn}, LSUN-C~\cite{yu2015lsun}, LSUN-R~\cite{yu2015lsun}, iSUN~\cite{Xu2015iSun}, Places365~\cite{zhou2017places}, DTD~\cite{cimpoi2014DTD}, TinyImageNet (TIN)~\cite{deng2009imagenet}, SUN~\cite{xiao2010sun}, and iNaturalist~\cite{van2018inaturalist}.
The CIFAR-10 or CIFAR-100 dataset was used as the ID dataset, and the other nine datasets were used as OOD datasets.

\noindent \textbf{Evaluation Measure.}
We used FPR95 as the primary evaluation measure, which is the FPR of OOD samples when the TPR for ID samples is 95.0\%.
ROC curves and the area under the curve (AUC) are also reported.

\noindent \textbf{Baselines.}
We chose five baseline methods:
MSP scoring \cite{hendrycks2017msp},
energy-based detection (Energy)~\cite{liu2020energy},
rectified activations applied to energy (ReAct)~\cite{sun2021react},
MHE~\cite{zhang23she}, and SHE~\cite{zhang23she}.
Note that the last four methods are energy-based OOD detection methods, with MHE and SHE being state-of-the-art using MHNs.
Also note that these methods, including ours, process representations from a frozen encoder. 
For a fair comparison, we use the same encoder in each experiment.
Another type of ODD detection methods (such as \citet{zhang23she}) that jointly optimize encoder and OOD module in a specific fashion is excluded.

\begin{figure*}
\centering
\includegraphics[width=\linewidth]{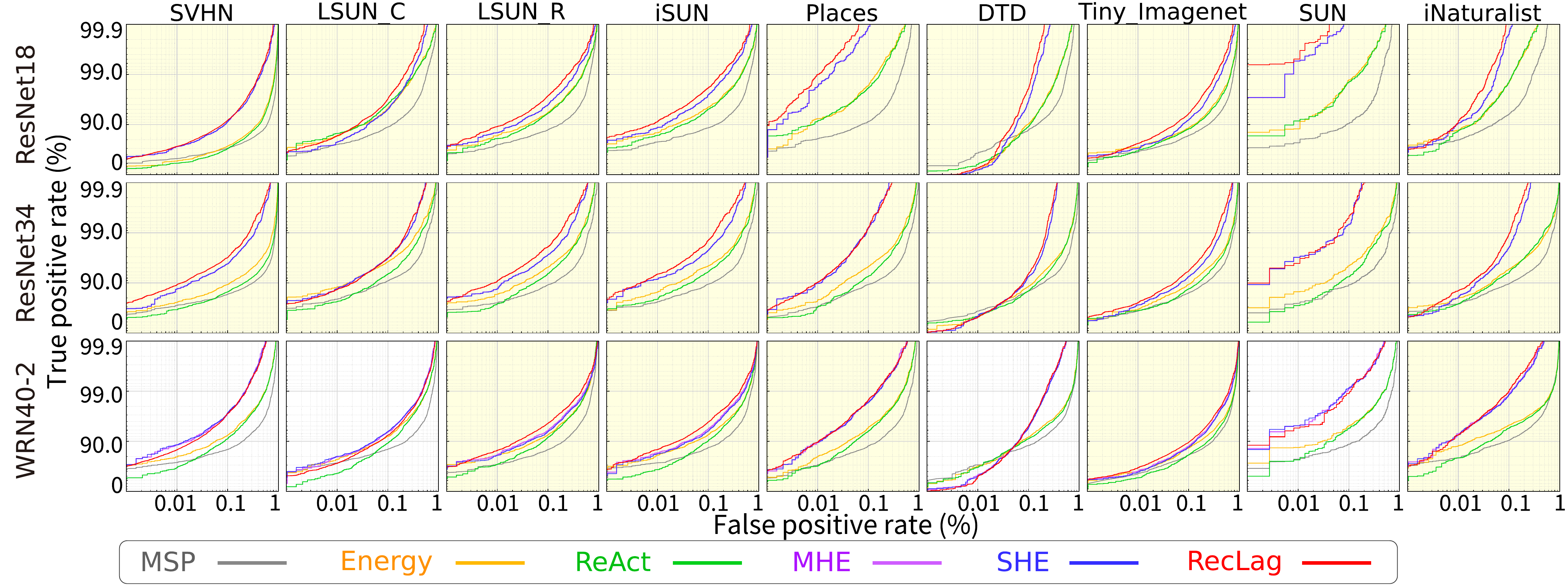}
\caption{ROC curves (log-scale) and AUC$\uparrow$. Yellow background indicates that RecLag performed the best in terms of AUC.}
\label{fig:roc}
\end{figure*}
\noindent \textbf{Neural networks.}
Three image classification networks were used: ResNet18~\cite{he2016resnet},
ResNet34~\cite{he2016resnet}, and
WideResNet40-2 (WRN40-2)~\cite{zagoruyko2016wide}.
They were trained on an ID dataset using cross-entropy loss. The OOD benchmark was conducted with no dynamics simulations (no extra computational costs compared to the baselines). Other implementation details are given in Appendix D.

\subsection{Experimental Results}
\noindent \textbf{Comparison With Energy-Based Methods.}
Table~\ref{tab:fpr} shows the OOD detection performance on the nine OOD datasets with the three neural networks trained on the CIFAR-10 dataset.
As shown, our RecLag-based MHN (RecLag) achieved the best average performance across all neural networks.
This demonstrates the effectiveness of our approach, which incorporates an attractor for OOD samples in post-hoc OOD detection scenarios.

\noindent \textbf{ROC Curves.}
Figure~\ref{fig:roc} reports the ROC curves with the AUC values. As shown, RecLag exhibited the best AUC value in 23 out of 27 comparisons (highlighted with the yellow background), indicating its consistent superiority in OOD detection performance.

\begin{table*}[t]
\centering
\setlength{\tabcolsep}{8.5pt}
{\footnotesize
\begin{tabular}{l|rrrrrrrrr|r}
\toprule
Method & SVHN & LSUN-C & LSUN-R & iSUN & Places & DTD & TIN & SUN & iNaturalist & Average \\
\midrule
MSP & 73.51 & 67.42 & 88.65 & 86.73 & 66.61 & 81.79 & 83.06 & 73.57 & 72.27 & 77.07\\
Energy & 66.00 & 54.96 & 82.88 & 82.23 & 56.55 & 78.85 & 77.49 & 66.21 & 70.86 & 70.67\\
ReAct & 61.33 & 52.73 & 83.36 & 83.48 & 53.61 & 74.92 & 77.27 & 62.67 & 66.29 & 68.41\\
MHE & 16.24 & 41.21 & 67.61 & 56.08 & 9.99 & 40.80 & 61.79 & 10.35 & 17.22 & 35.70\\
SHE & 16.15 & 41.07 & 67.78 & 56.42 & \textbf{9.91} & 40.37 & 61.89 & \textbf{10.08} & \textbf{16.90} & 35.61\\
RecLag & \textbf{15.50} & \textbf{39.94} & \textbf{65.28} & \textbf{55.67} & 11.57 & \textbf{39.18} & \textbf{59.02} & 12.17 & 19.29 & \textbf{35.29} \\
& \textbf{$\pm$ 3.09} & \textbf{$\pm$ 0.80} & \textbf{$\pm$ 3.96} & \textbf{$\pm$ 4.52} & $\pm$ 0.54 & \textbf{$\pm$ 1.13} & \textbf{$\pm$ 4.43} & $\pm$ 0.64 & $\pm$ 1.29 & \textbf{$\pm$ 1.93} \\
\bottomrule
\end{tabular}
}
\caption{OOD detection performance as
FPR95(\%) $\downarrow$ with CIFAR-100 images being ID samples.
WRN40-2 arch.~was used.
For other descriptions, see the caption of Table~\ref{tab:fpr}.
}
\label{tab:c100}
\end{table*}
\begin{figure}
\centering
\includegraphics[width=\linewidth]{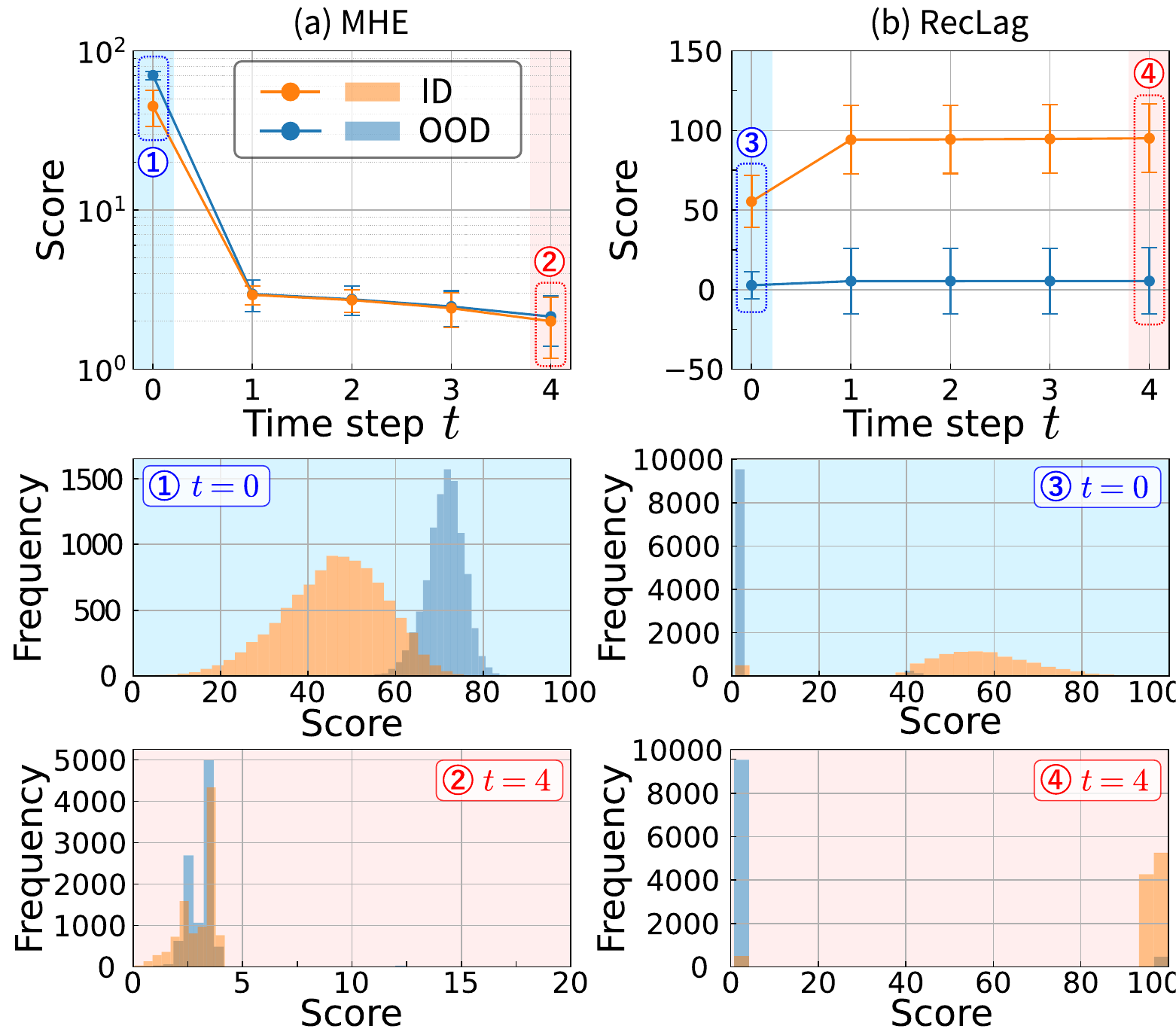}
\caption{Comparison detection scores over time on LSUN-R. ResNet18 trained on CIFAR-10 was used.}
\label{fig:score}
\vspace{14pt}
\centering
\includegraphics[width=0.95\linewidth]{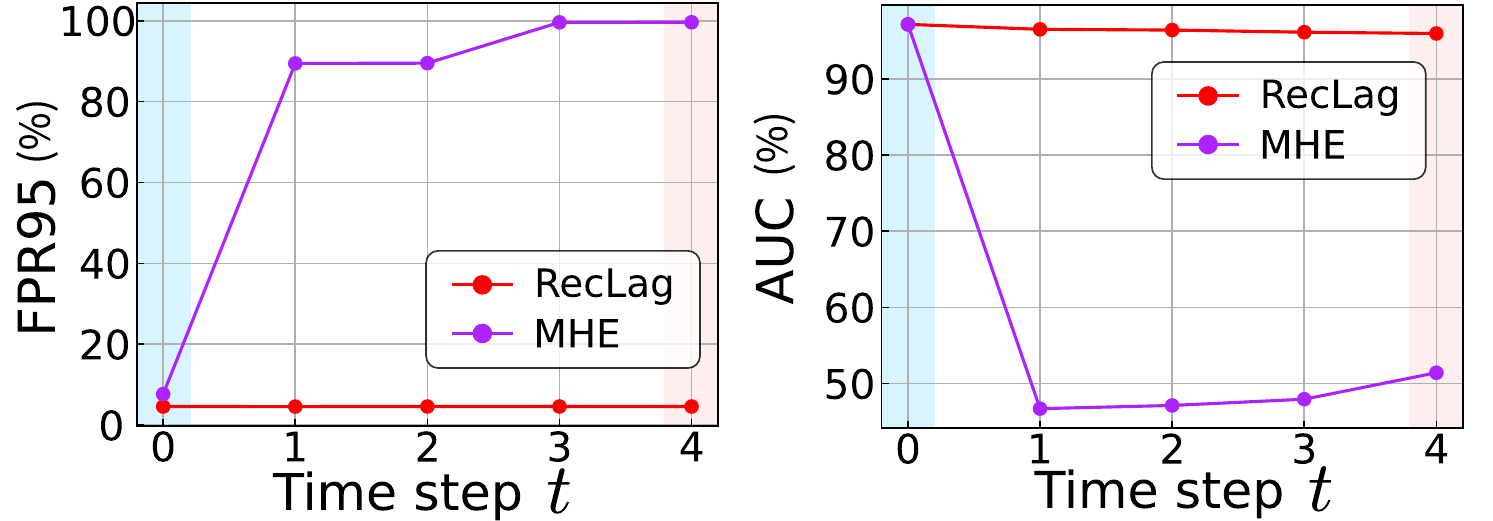}
\caption{FPR95 $\downarrow$ and AUC $\uparrow$ over time on LSUN-R.}
\label{fig:auc}
\end{figure}

\noindent \textbf{In-Distribution Data.}
To investigate how OOD detection performance is affected when a neural network is trained on a more complex task, Table~\ref{tab:c100} shows the results for WRN40-2 trained on CIFAR-100.
As shown, the OOD performance decreases for all methods compared to those in Table 1. This is because the variance of features in ID samples increased, making OOD detection more challenging.
However, even in this case, our RecLag-based MHN outperformed the other methods. This result indicates that the relative effectiveness of our approach is robust against differences in ID samples.

\noindent \textbf{Time Evolution.}
Figure~\ref{fig:score} analyzes how the detection scores change as the dynamical system of MHNs evolves over time.
With MHE, OOD samples have higher energy scores than ID samples at time $t = 0$; however, the scores decrease over time, making it almost impossible to distinguish between ID samples and OOD samples at a discrete time step of 4.
In contrast, RecLag-based MHN can distinguish between ID samples and OOD samples even after the score converges, thereby maintaining OOD detection performance over time as shown in Figure~\ref{fig:auc}.
This demonstrates that our approach successfully managed OOD samples within the dynamical system of MHNs.

\begin{figure}
    \centering
\includegraphics[width=0.9\linewidth]{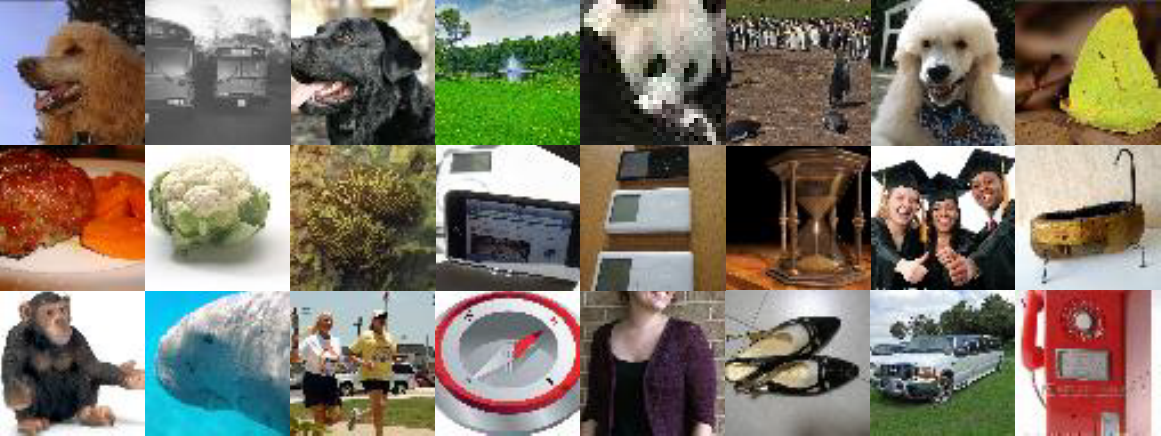}
\caption{OOD samples incorrectly identified as ID samples. ID: CIFAR-100, OOD: TIN.}
\label{fig:analysis}
\end{figure}

\noindent \textbf{Visual Analysis.}
Figure~\ref{fig:analysis} analyzes the OOD samples incorrectly identified as ID samples.
As shown, images of animals, people, and foods were difficult to detect as OOD. This study focused on post-hoc OOD detection, but in the future, it would be interesting to simultaneously train MHNs and classification networks to further improve the performance.

%% file: 7_conslusion.tex
\section{Conclusion}
We proposed the RecLag function, a specially designed Lagrangian
to equip MHNs with OOD rejection functionality.
In our method, the interaction matrix is optimized so as to compute probability densities, which are used to determine ID/OOD.
Theoretically, RecLag-based MHNs reduces to vanilla MHNs when the ID memory strength is infinitely large; therefore, the proposed method is a natural extension of existing MHNs.
Experiments on nine image datasets demonstrated the effectiveness of our approach, surpassing energy-based OOD detection methods. 

\noindent \textbf{Limitation.}
While this work introduced a new Lagrangian for memory neurons, the Lagrangian for feature neurons remains underexplored. Similar to previous works, we used the activation function $g$ that takes the simplest form in Euclidean space, $g_{i}(v) =v_{i}$, because recent deep learning efforts often assume that the feature space is Euclidean.
Investigating new Lagrangians in other non-linear feature spaces, such as spherical or hyperbolic space, might be promising.

\noindent \textbf{Future Work.}
In future work, we will focus on generalizing RecLag for structured memory patterns such as hierarchical memory patterns. Applications to regression tasks are also intriguing.
We believe this work has opened up new avenues for exploring the potential of MHNs.

\section*{Acknowledgements}
This work was supported by JSPS KAKENHI Grant Number JP22H03642 and DENSO IT LAB Recognition and Learning Algorithm Collaborative Research Chair (Science Tokyo).
We thank Toshihiro Ota for fruitful discussion.

%% file: 8_appendix.tex
\clearpage
\onecolumn
\section*{Appendix A. Proof of Theorem 1}

Theorem 1 shows that a point attractor exists at the origin of the feature space.
We first define the attracting set $A$ and its basin $B$, and then provide a proof of Theorem 1.

\vspace{5pt}
\noindent \textit{\textbf{Definition 3.}
Let $\mathcal{X}$ be a Hausdorff space
and $\varphi(t, x)$ be a dynamical system on $\mathcal{X}$, where $t \in \mathbb{N}$ is a time index, $x \in \mathcal{X}$ is an initial point, and all $\varphi(t, \cdot): \mathcal{X} \to \mathcal{X}$ are continuous functions.
We say that a closed set $A \subset \mathcal{X}$ is an attracting set if there exists a neighborhood $U$ of $A$ that satisfies the following two conditions.
{
\setlength{\leftmargini}{20pt}
\begin{enumerate}
\item [(a)] There exists $T$ such that ~$\bigcap_{t \ge T}\{\varphi(t, x) : x \in U\} = A$.
\item [(b)]
There exists $T$ such that,
for every neighborhood $V$ of $A$,~$t \ge T \Rightarrow\{\varphi(t, x) : x \in U\} \subset V$.
\end{enumerate}
}
\noindent We define the basin $B$ of attraction of $A$ as $B = \bigcup_{t \ge 0} \{x : \varphi(t, x) \in U\}$.
}

\vspace{5pt}
\noindent \textit{\textbf{Theorem 1.}
Suppose that activation functions $f$ and $g$ in the dynamical system of Eqs.~(\ref{eq:de_v}, \ref{eq:de_h}) are given by the derivatives of RecLag $\Lagh$ in Eq.~(\ref{eq:reclag}) and the Lagrangian $\Lagv$ in Eq.~(\ref{eq:lag_modern}), respectively.
For any interaction matrix $\xi \in \mathbb{R}^{\Numh \times \Numv}$, a trivial point attracting set $A = \{\bm{0}\}$ exists at the origin
$\bm{0} \in \mathbb{R}^{\Numv}$ in the feature space when $\gamma > \Numh$ under the adiabatic limit $\tauv = dt$.}

\renewcommand{\proofname}{Proof}
\begin{proof}
With RecLag, the activation function $f_{\nu}$ is given by
\begin{align}
f_{\nu}(h)
&= \frac{\partial}{\partial h_{\nu}} \max \left(\log \left( \frac{1}{\gamma} \sum^{\Numh}_{\mu=1} \exp \left({\beta h_{\mu}}\right) \right)^{\frac{1}{\beta}}, 0 \right)\\
&= \chi \left( \log \left( \frac{1}{\gamma} \sum^{\Numh}_{\mu=1} \exp \left({\beta h_{\mu}}\right) \right)^{\frac{1}{\beta}} \right) \cdot \frac{\partial}{\partial h_{\nu}} \log \left( \frac{1}{\gamma} \sum^{\Numh}_{\mu=1} \exp \left({\beta h_{\mu}}\right) \right)^{\frac{1}{\beta}}\\
&= 
\chi \left(
\frac{1}{\beta}
\log \left( \frac{1}{\gamma} \sum^{\Numh}_{\mu=1} \exp \left({\beta h_{\mu}}\right) \right)
\right)
\cdot
\frac{1}{\beta}
\left( \frac{1}{\gamma} \sum^{\Numh}_{\mu=1} \exp \left({\beta h_{\mu}}\right) \right)^{-1} \cdot \frac{\beta}{\gamma} \exp \left({\beta h_{\nu}}\right)\\
&= 
\chi \left(
\log \left( \frac{1}{\gamma} \sum^{\Numh}_{\mu=1} \exp \left({\beta h_{\mu}}\right) \right)
\right)
\cdot
\left( \sum^{\Numh}_{\mu=1} \exp \left({\beta h_{\mu}}\right) \right)^{-1} \cdot  \exp \left({\beta h_{\nu}}\right)\\
&= 
\chi \left(
\log \left( \frac{1}{\gamma} \sum^{\Numh}_{\mu=1} \exp \left({\beta h_{\mu}}\right) \right)
\right)
\cdot
\mathrm{softmax}_{\nu} (\beta h) \label{eq:23}
\end{align}
where
\begin{align}
\chi (x) =
\begin{cases}
1 & (x \geq 0)\\
0 & (x < 0)
\end{cases}.
\end{align}
Under the adiabatic limit, {\it i.e.}, when the dynamics of memory neurons is changing rapidly, we have
\begin{align}
h_{\mu} = \sum_{j=1}^{\Numv} 
\xi_{\mu j} v_{j}.
\end{align}
Thus, we obtain
\begin{align}
\text{Eq.~(\ref{eq:23})}
&=
\chi \left(
\log \left( \frac{1}{\gamma} \sum^{\Numh}_{\mu=1} \exp \left({\beta \sum_{j=1}^{\Numv} 
\xi_{\mu j} v_{j}}\right) \right)
\right)
\cdot
\mathrm{softmax}_{\nu} \left(\beta \sum_{j=1}^{\Numv} 
\xi_{\cdot j} v_{j} \right)\\
&= \chi(G(v)) \cdot
\mathrm{softmax}_{\nu} \left(\beta \sum_{j=1}^{\Numv} 
\xi_{\cdot j} v_{j} \right).
\end{align}
where
\begin{align}
G(v) =
\log
\left(
\frac{1}{\gamma}
\sum_{\mu = 1}^{\Numh}
\exp\left({\beta \sum_{j=1}^{\Numv} \xi_{\mu j} v_{j}}\right)
\right).
\end{align}
The differential equation in Eq.~(\ref{eq:de_v}) is then written by
\begin{align}
\label{eq:29}
\tauv \frac{d v_{i}(t)}{d t} &= \sum_{\mu=1}^{\Numh} \xi_{i \mu} f_{\mu}(h(t)) - v_{i}(t)\\
&= \sum_{\mu=1}^{\Numh} \xi_{i \mu} \chi(G(v(t)))
~\mathrm{softmax}_{\mu} \left(\beta \sum_{j=1}^{\Numv} 
\xi_{\cdot j} v_{j}(t) \right) - v_{i}(t)\\
&= \chi(G(v(t))) \sum_{\mu=1}^{\Numh} \xi_{i \mu} ~\mathrm{softmax}_{\mu} \left(\beta \sum_{j=1}^{\Numv} 
\xi_{\cdot j} v_{j}(t) \right) - v_{i}(t).\label{eq:31}
\end{align}
To derive the update rule, we consider the first-order Taylor approximation
\begin{align}
v_{i}(t + \Delta t)
=
v_{i}(t) + \frac{d v_{i}(t)}{d t} \Delta t,
\end{align}
where $\Delta t$ is a small time step.
From Eq.~(\ref{eq:31}), we have 
\begin{align}
v_{i}(t + \Delta t)
&=
v_{i}(t)
+
\frac{\Delta t}{\tauv}
\left(
\chi(G(v(t))) \sum_{\mu=1}^{\Numh} \xi_{i \mu} ~\mathrm{softmax}_{\mu} \left(\beta \sum_{j=1}^{\Numv} 
\xi_{\cdot j} v_{j}(t) \right) - v_{i}(t)
\right).
\end{align}
Therefore, when $\tauv = \Delta t$, we have
\begin{align}
v_{i}(t + \Delta t)
&=
\chi(G(v(t))) \sum_{\mu=1}^{\Numh} \xi_{i \mu} ~\mathrm{softmax}_{\mu} \left(\beta \sum_{j=1}^{\Numv} 
\xi_{\cdot j} v_{j}(t) \right)
.
\end{align}
This yields the update rule with discrete time steps $k \in \mathbb{N}$ as follows:
\begin{align}
v^{(k+1)}_{i}
=
\chi
\left(
G(v^{(k)})
\right)
\sum_{\mu=1}^{\Numh}
\xi_{i\mu}
~\mathrm{softmax}_{\mu}
\left(
\beta \sum_{j=1}^{\Numv} 
\xi_{\cdot j} v^{(k)}_{j}
\right).
\end{align}

Finally, we show that $A = \{\bm{0}\}$ is an attracting set for every fixed $\xi$.
Suppose that $\mathcal{X} = \mathbb{R}^{\Numv}$ is the feature space.
We consider the Euclidean distance $d(x, y) = \|x - x^{\prime}\|_{2}$ between two points $x, x^{\prime} \in \mathcal{X}$.
Clearly, with the topology induced by the open balls
\begin{align}
U_{\epsilon}(x) = \{ x^{\prime} \in \mathcal{X} : d(x, x^{\prime}) < \epsilon\}~~(\epsilon > 0),
\end{align}
the space $\mathcal{X}$ is a Hausdorff space. The dynamic system $\varphi(k, x)$ is then given by
\begin{align}
\varphi(k, x) =
\begin{cases}
x & (k = 0)\\
\chi
\left(
G(\varphi(k, x))
\right)
\sum_{\mu=1}^{\Numh}
\xi_{i\mu}
~\mathrm{softmax}_{\mu}
\left(
\beta \sum_{j=1}^{\Numv} 
\xi_{\cdot j} \varphi_{j}(k, x)
\right) & (k > 0)
\end{cases}.
\end{align}
Below, we show that the two conditions (a) and (b) in Definition 3 are satisfied.

\noindent \textit{Proof of (a).}
Let $U = U_{\epsilon}(\bm{0})$ be an open ball with
\begin{align}
\label{eq:38}
\epsilon = \frac{1}{\Numv \beta~\Xi} \log \frac{\gamma}{\Numh},
\quad
\Xi = \max_{\mu, j} |\xi_{\mu j}|,
\end{align}
where $\gamma > \Numh$.
For every $x \in U$, we have
\begin{align}
G(\varphi (0, x))
&=
\log
\left(
\frac{1}{\gamma}
\sum_{\mu = 1}^{\Numh}
\exp\left({\beta \sum_{j=1}^{\Numv} \xi_{\mu j} x_{j}}\right)
\right)\\
&=
\log \left(
\sum_{\mu = 1}^{\Numh}
\exp\left({\beta \sum_{j=1}^{\Numv} \xi_{\mu j} x_{j}}\right)
\right)
- \log \gamma\\
&\leq
\log \left(
\Numh
\max_{\mu} \exp \left( {\beta \sum_{j=1}^{\Numv} \xi_{\mu j} x_{j}}\right)\right)
- \log \gamma\\
&\leq
\max_{\mu} \left({\beta \sum_{j=1}^{\Numv} \xi_{\mu j} x_{j}}
\right) - \log \frac{\gamma}{\Numh}\\
&\leq
\Numv \beta \max_{\mu} \max_{j} \left({\xi_{\mu j} x_{j}}
\right) - \log \frac{\gamma}{\Numh}\\
& \leq
\Numv \beta~\Xi~\|x\|_{\infty}
 - \log \frac{\gamma}{\Numh}\\
 & \leq
\Numv \beta~\Xi~\|x\|_{2}
 - \log \frac{\gamma}{\Numh}\\
& <
\Numv \beta~\Xi~\epsilon
 - \log \frac{\gamma}{\Numh}\\
&= 0.
\end{align}
Thus, when $T = 1$, we have
\begin{align}
\bigcap_{t \ge T} \{\varphi(t, x) : x \in U_{\epsilon}\}
= \bigcap_{t \ge T} \{\bm{0}\} =
A.
\end{align}

\noindent \textit{Proof of (b).}
Suppose that $V = \{x : d(x, \bm{0}) < \epsilon^{\prime}\}$ is a neighborhood of $A$.
With the open ball $U = U_{\epsilon}(\bm{0})$ defined by Eq.~(\ref{eq:38}) and when $T = 1$, we have
\begin{align}
\{\varphi(t, x) : x \in U\} = A \subset V,
\end{align}
when $t \geq T$.

This shows that $A$ is an attracting set when $\gamma > \Numh$ for every fixed $\xi$.
\end{proof}

\section*{Appendix B. Proof of Theorem 2}
\textit{\textbf{Theorem 2.}
Let $v_{\mathrm{A}}$ and $v_{\mathrm{B}}$ be feature neurons of a vanilla MHN and a RecLag-based MHN, respectively. Suppose that $v_{\mathrm{A}}^{(0)} = v_{\mathrm{B}}^{(0)}$. For every $\epsilon > 0$,
a small $\gamma > 0$ exists such that $\sup_{k} \|v_{\mathrm{A}}^{(k)} - v_{\mathrm{B}}^{(k)}\|_{2} < \epsilon$.
}

\begin{proof}
The update rules for $v_{\mathrm{A}}$ and $v_{\mathrm{B}}$ are given by
\begin{align}
v^{(k+1)}_{\mathrm{M},i}
&=
\sum_{\mu=1}^{\Numh}
\xi_{i\mu}
~\mathrm{softmax}_{\mu}
\left(
\beta \sum_{j=1}^{\Numv} 
\xi_{\cdot j} v^{(k)}_{\mathrm{M},j}
\right),\\
v^{(k+1)}_{\mathrm{R},i}
&=
\chi
\left(
G(v^{(k)}_{\mathrm{B}})
\right)
\sum_{\mu=1}^{\Numh}
\xi_{i\mu}
~\mathrm{softmax}_{\mu}
\left(
\beta \sum_{j=1}^{\Numv} 
\xi_{\cdot j} v^{(k)}_{\mathrm{R},j}
\right).
\end{align}
Let $0 < \delta < 1$ be a small constant and
\begin{align}
\gamma = \delta \min_{k} \sum_{\mu = 1}^{\Numh}
\exp\left({\beta \sum_{j=1}^{\Numv} \xi_{\mu j} v^{(k)}_{\mathrm{M},j}}\right).
\end{align}
When $v^{(k)}_{\mathrm{B}} = v^{(k)}_{\mathrm{A}}$, we have
\begin{align}
\chi(G(v^{(k)}_{\mathrm{B}}))
&=
\chi
\left(
\log
\left(
\frac{1}{\gamma}
\sum_{\mu = 1}^{\Numh}
\exp\left({\beta \sum_{j=1}^{\Numv} \xi_{\mu j} v^{(k)}_{\mathrm{R},j}}\right)
\right)
\right)\\
&=
\chi
\left(
\log
\left(
\sum_{\mu = 1}^{\Numh}
\exp\left({\beta \sum_{j=1}^{\Numv} \xi_{\mu j} v^{(k)}_{\mathrm{M},j}}\right)
\right)
- \log{\gamma}
\right)\\
&= \chi
\left(
\log
\frac{
\sum_{\mu = 1}^{\Numh}
\exp\left({\beta \sum_{j=1}^{\Numv} \xi_{\mu j} v^{(k)}_{\mathrm{M},j}}\right)
}{
\min_{k} \sum_{\mu = 1}^{\Numh}
\exp\left({\beta \sum_{j=1}^{\Numv} \xi_{\mu j} v^{(k)}_{\mathrm{M},j}}\right)
}
- \log \delta\right)\\
&= 1
,
\end{align}
and we have
\begin{align}
\|v_{\mathrm{A}}^{(k+1)} - v_{\mathrm{B}}^{(k+1)}\|^{2}_{2}
&=
\sum_{i=1}^{\Numv}
\left(
\left( 1 - \chi
\left(
G(v^{(k)}_{\mathrm{B}})
\right)
\right)
\sum_{\mu=1}^{\Numh}
\xi_{i\mu}
~\mathrm{softmax}_{\mu}
\left(
\beta \sum_{j=1}^{\Numv} 
\xi_{\cdot j} v^{(k)}_{\mathrm{M},j}
\right)
\right)^{2}\\
&=
\left( 1 - \chi
\left(
G(v^{(k)}_{\mathrm{B}})
\right)
\right)^{2}
\sum_{i=1}^{\Numv}
\left(
\sum_{\mu=1}^{\Numh}
\xi_{i\mu}
~\mathrm{softmax}_{\mu}
\left(
\beta \sum_{j=1}^{\Numv} 
\xi_{\cdot j} v^{(k)}_{\mathrm{M},j}
\right)
\right)^{2}\\
&= 0.
\end{align}
This assumption gives us $v^{(0)}_{\mathrm{B}} = v^{(0)}_{\mathrm{A}}$, and thus, for every $\epsilon > 0$,
\begin{align}
\sup_{k} \|v_{\mathrm{A}}^{(k)} - v_{\mathrm{B}}^{(k)}\|_{2} = 0< \epsilon
\end{align}
\end{proof}

\section*{Appendix C. Proof of Theorem 3}
\textit{\textbf{Theorem 3.}
The basin $B_{0} = \{ v : G(v) < 0\}$ is identical to the set of points that have low probability density values, {\it i.e.}, a threshold $\delta$ exists such that
\begin{align}
B_{0} = \{x : p_{H}(X = x) < \delta\}.
\end{align}
}
\begin{proof}
With the joint probability distribution $p_{H}(X=x, M=\mu)$ given by Definition~2, the marginal distribution $p_{H}(X=x)$ is given by
\begin{align}
p_{H}(X = x)
=
\sum_{\mu=1}^{\Numh}
\frac{1}{Z} \exp \left(
{\beta \sum_{j=1}^{\Numv} \xi_{\mu j} x_{j}}
\right)
=
\frac{\gamma}{Z} \exp(G(x)).
\end{align}
Therefore, for fixed values of $\xi$ and $\gamma$, we have
\begin{align}
\delta = \frac{\gamma}{Z}
\end{align}
satisfying
\begin{align}
B_{0}
= \{v: G(v) < 0\}
= \{v: \exp(G(v)) < 1\}
= \left\{x: p_{H}(X = x) < \delta \right\}.
\end{align}
This shows that the basin is a set of data samples that have a probability density lower than $\delta$.
\end{proof}

\section*{Appendix D. Implementation details}

Three image classification networks were used: ResNet-18~\cite{he2016resnet},
ResNet-34~\cite{he2016resnet}, and
WideResNet40-2~\cite{zagoruyko2016wide}.
Each network was trained on an ID dataset using cross-entropy loss for 200 epochs with an SGD momentum optimizer. The initial learning rate was set to 0.1, and it was decayed by a factor of 0.1 at 100 and 150 epochs. The batch size was set to 128. Random cropping and horizontal flipping were used to augment the training images.
The dimension of the output representation was 512, and thus the number of feature neurons was set as $N_{V} = 512$.
During the training of the interaction matrix, normalization was applied to the output representations so that the L2 norms are 10.0.
The interaction matrix was trained for 100 epochs by following the training method for SFNN proposed by \citet{tang13sfnn}, where the number of memory neurons is set as $N_{H} = 250$, the inverse temperature parameter $\beta$ is set to 5.0, and the number of samples for Monte Carlo approximation is set to 5.
The objective function was computed using the input features as targets, as described in Eq.~(\ref{eq:objective_sfnn}). The OOD datasets were prepared following \citet{shen2023posthoc}, with all images resized to $32 \times 32$. We used the official implementation of Energy~\cite{liu2020energy}, ReAct~\cite{sun2021react}, MHE and SHE~\cite{shen2023posthoc} to report their results.